\documentclass[conference]{IEEEtran}

\IEEEoverridecommandlockouts

\usepackage{amsmath,amssymb,amsfonts}
\usepackage{graphicx}
\usepackage{booktabs}
\usepackage{multirow}
\usepackage{array}
\usepackage{xcolor}
\usepackage{enumitem}
\usepackage{cite}
\usepackage{url}
\usepackage[hidelinks]{hyperref}

\begin{document}

% =================================================================
%  Title
% =================================================================
\title{Discrete Action Space as a Prerequisite for GRPO Convergence in Small-Model Continuous Control}

% --- Authors ------------------------------------------------------
% IEEE 6-author template. Replace placeholders with real names,
% affiliations, and emails before submission.
% ------------------------------------------------------------------
\author{
    \IEEEauthorblockN{Dmytro Filatov}
    \IEEEauthorblockA{\textit{Applied AI and Computer Vision} \\
    \textit{Aimech Technologies Corp.}\\
    San Francisco, The United States of America \\
    dima@deepxhub.com}
    \and
    \IEEEauthorblockN{Valentyn Fedorov}
    \IEEEauthorblockA{\textit{Artificial Intelligence and Computer Vision} \\
    \textit{Aimech Technologies Corp.}\\
    Kyiv, Ukraine \\
    valentyn.fedorov@deepxhub.com}
    \and
    \IEEEauthorblockN{Vira Filatova}
    \IEEEauthorblockA{\textit{Applied Artificial Intelligence} \\
    \textit{Covijn Ltd.}\\
    London, United Kingdom \\
    vira@deepxhub.com}
}

\maketitle

% =================================================================
%  Abstract
% =================================================================
\begin{abstract}
Discrete action space is a structural prerequisite, not a stylistic
preference, for Group Relative Policy Optimization (GRPO) to
converge when fine-tuning small ($\le 1$\,B) language models in
continuous-control settings. Vanilla GRPO fine-tuning of Qwen-0.5B
for 25\,Hz quadrotor velocity control collapses to the trivial
zero action: 0\,\% success rate, entropy falling from 0.35 to 0.03
within 60 steps. Two ablations -- removing the jerk-penalty term,
and removing the KL anchor to the pretrained prior -- each prevent
entropy collapse, yet neither enables learning. Only when the
action space is replaced by a 5-way categorical choice over PID
presets does training converge. The resulting controller traces a
smoothness--reliability Pareto frontier along training duration;
both endpoints are reported (98.6\,\% / 0.656\,m/s$^3$ at 64
steps; 100\,\% / 1.103, or 0.796 under matched velocity-cap, at
256 steps). The recipe replicates across three pretrained language
model families. As context, a re-tuned classical baseline (PID
with $K_i\!=\!0.30$, $v_{\max}\!=\!2.5$) reaches the same 100\,\%
success rate at jerk 0.736\,m/s$^3$. A high-fidelity simulation
using Crazy{f}lie 2.1 dynamics surfaces a hover-region training-
distribution gap.

\end{abstract}

% =================================================================
\section{Introduction}

Small language models (SLMs, $\le 1$\,B parameters) fit in under
one gigabyte of GPU memory, run at tens of hertz on edge
accelerators, and inherit broad linguistic priors from pretraining.
That makes them an attractive substrate for on-board UAV mission
planning: a fine-tuned SLM could in principle bridge
language-conditioned intent to closed-loop actuation without the
7B--70B parameter footprint of current vision-language-action
systems (RT-2~\cite{rt2}, OpenVLA~\cite{openvla},
$\pi_0$~\cite{pi0}).

The standard route for turning a language model into a specialised
agent is reinforcement-learning fine-tuning. Group Relative Policy
Optimization (GRPO)~\cite{deepseekmath} has become the workhorse of
chain-of-thought alignment: it needs only a verifiable reward and
no separate value network. The natural question is whether the same
recipe -- plug in a dense reward, turn the crank -- transfers to
continuous control, where the ``reasoning tokens'' are numerical
velocities rather than English.

It does not, in our setting, and the reason has nothing to do with
reward design or regularisation. The reason is the structure of the
action space itself. Four ablations support this:

\begin{enumerate}
\item A 21-scenario quadrotor benchmark spans six disturbance
  families, evaluated at $N\!=\!10$ seeds with bootstrap and
  Wilson 95\,\% confidence intervals; the benchmark is released as
  open source.

\item Among four GRPO configurations -- vanilla,
  no-jerk-penalty, no-KL-anchor, and discrete-categorical -- only
  the discrete-categorical run produces a learning signal. The
  other three either collapse to a fixed action or stall in the
  noise floor.

\item The resulting controller traces a smoothness--reliability
  Pareto frontier whose endpoints we report jointly. A
  velocity-cap confound is resolved post-hoc, and the recipe is
  cross-validated on three pretrained LM families against a
  re-tuned classical PID baseline that matches the trained
  controller's reliability without the LM.
\end{enumerate}

% =================================================================
\section{Related Work}
\subsection{RL Fine-tuning of Language Models}

PPO-based RLHF~\cite{stiennon2020,ouyang2022},
DPO~\cite{rafailov2023}, and GRPO~\cite{deepseekmath} form the
canonical RL alignment recipe family. Prior GRPO work focuses on
discrete symbolic reasoning (math, code, tool use), where the base
model already places mass on plausible completions and GRPO
\textit{refines} rather than \textit{elicits}. Our work extends GRPO
into continuous control, a setting it was not originally designed
for, and identifies the action-space structure that controls whether
convergence is possible at all.

\subsection{Foundation Models for Robotic Control}

End-to-end VLAs (RT-2~\cite{rt2}, OpenVLA~\cite{openvla},
$\pi_0$~\cite{pi0}) are 3B--70B-parameter end-to-end controllers
that do not fit on-board SWaP-constrained UAV compute. Hierarchical
patterns (Code-as-Policies~\cite{liang2023}, SayCan~\cite{ahn2022},
Inner Monologue~\cite{huang2022}) place a language model at the
planning level and delegate inner control to classical methods. Our
5-way categorical mode selector is an instance of this hierarchy
with the action-space parameter set to a discrete categorical at
2.5\,Hz.

\subsection{Classical Quadrotor Control}

PID~\cite{bouabdallah2007}, MPC~\cite{kamel2017}, and adaptive
variants~\cite{antonelli2018,bisheban2021} provide mature baselines.
We re-tune the PID baseline in Section~\ref{ssec:pid_baseline} and
find that, contrary to common framing in the literature, a single
high-integral-gain PID suffices for our benchmark.

\subsection{Mode Collapse and Reward Hacking}

Entropy regularisation~\cite{mnih2016}, MaxEnt
RL~\cite{haarnoja2018}, and RLHF
over-optimisation~\cite{gao2023} have been studied in
discrete-token settings. Our analysis extends to the
continuous-control setting, with controlled mechanism probes that
show no single named factor (jerk-penalty, KL-anchor) explains the
failure alone.

% =================================================================
\section{Methodology}
\subsection{Environment and Benchmark}

A 1.5\,kg quadrotor in a first-order velocity-tracking model with
rotor time constant $\tau\!=\!0.5$\,s. Control runs at 25\,Hz
($dt\!=\!0.04$\,s) in the local NED frame; per-axis velocity
commands are clipped to $[-5,5]$\,m/s. Each scenario draws from a
configurable disturbance suite:

\begin{itemize}
\item \textbf{Wind:} Dryden turbulence
  model~\cite{dryden1943,milf8785c} with mean speed
  $w \in [0, 8]$\,m/s and adjustable turbulence intensity.
\item \textbf{Payload:} pickup with abrupt mass change
  $m_p \in \{0.3, 0.5, 1.0, 1.5\}$\,kg, optionally mid-flight.
\item \textbf{Sensor noise:} Gaussian on position and velocity
  observations.
\item \textbf{Actuation:} motor degradation factor and FIFO
  communication delay.
\end{itemize}

The 21-scenario benchmark covers six families
(Table~\ref{tab:scenarios}). Full disturbance parameters are
released with the source code.

\begin{table}[!t]
\caption{The 21-scenario benchmark.}
\label{tab:scenarios}
\centering
\renewcommand{\arraystretch}{1.1}
\footnotesize
\begin{tabular}{ll}
\toprule
\textbf{Family} & \textbf{Scenarios} \\
\midrule
Standard (6) & straight 5\,m, straight 10\,m, diagonal 15\,m, \\
             & payload 0.5\,kg, wind 3\,m/s, compound \\
Far distance (3) & 25\,m, 50\,m diagonal, 100\,m straight \\
Strong wind (3) & 5\,m/s, 8\,m/s, 8\,m/s + gusty \\
Heavy payload (3) & 1.0\,kg, 1.5\,kg, mid-flight pickup \\
Multi-waypoint (2) & 4-waypoint patrol, 6-waypoint zigzag \\
Multi-stressor (4) & wind+payload-far, wind+gust+payload, \\
                    & triple stressor, extreme triple \\
\bottomrule
\end{tabular}
\end{table}

\subsection{The Four Training Configurations}

We design four GRPO training runs that share base model, LoRA
configuration, curriculum, and per-step compute, and differ by one
isolated factor. The base configuration is Qwen2.5-0.5B-Instruct
with LoRA $r\!=\!8$, $\alpha\!=\!32$, applied to $\{q,k,v,o\}_{\rm
proj}$. 4-stage curriculum (5\,m $\to$ 10\,m $\to$ 15\,m $\to$
compound). Group size 8, $\beta\!=\!0.04$ unless overridden,
temperature 0.7. Reward is a 7-term dense composition (progress,
distance, reach, closest-approach, jerk, alignment, idle penalty;
full formulas released with code).

\textit{1) Experiment A - Vanilla GRPO:} Continuous JSON velocity
output, 7-term reward, 252 steps. The reference baseline.

\textit{2) Experiment B - No jerk penalty:} Same as A but the
$r_{\rm jerk}$ term is removed; KL anchor retained. 64 steps.

\textit{3) Experiment C - No KL anchor:} Same as A but
$\beta\!=\!0$; full 7-term reward retained. 64 steps.

\textit{4) Experiment D - Categorical action space:} Same as A
but the action space is restricted to one of five labels:
$\{\textsf{cruise},\textsf{brake},\textsf{hover},
\textsf{aggressive},\textsf{anti\_wind}\}$. Each label maps to a
pre-tuned PID gain tuple (Table~\ref{tab:presets}); the PID runs
at 25\,Hz, the LM is queried every 15 control steps (2.5\,Hz). We
train two checkpoints: \textbf{D-64} (64 steps) and \textbf{D-256}
(256 steps with oversampled wind+payload curriculum stages).

\begin{table}[!t]
\caption{PID gain presets per categorical label.}
\label{tab:presets}
\centering
\renewcommand{\arraystretch}{1.1}
\footnotesize
\begin{tabular}{lcccc}
\toprule
Label & $K_p$ & $K_i$ & $K_d$ & $v_{\max}$ \\
\midrule
\textsf{cruise} & 1.2 & 0.05 & 0.8 & 3.5 \\
\textsf{brake} & 0.8 & 0.03 & 1.6 & 2.0 \\
\textsf{hover} & 0.4 & 0.00 & 0.4 & 0.3 \\
\textsf{aggressive} & 2.0 & 0.10 & 1.2 & 4.5 \\
\textsf{anti\_wind} & 1.5 & 0.30 & 1.0 & 2.5 \\
\bottomrule
\end{tabular}
\end{table}

A deterministic stuck-detector forces the \textsf{anti\_wind}
preset if the drone is within 2.5\,m of the target with speed
$<0.3$\,m/s and no progress for 3 consecutive steps. The detector
operates on the classical control layer and does not require the
language model.

\paragraph{Justification} The stuck condition (small residual
error under sustained wind and payload disturbance) lies below the
language model's level of abstraction. The prompt contains no
``stuck'' token; the integral component of the PID regulator that
recovers from this state is part of the classical control loop the
LM cannot reach. Changing the LM's label preference cannot modify
the integrator's history. We therefore treat the stuck detector
on par with anti-windup clamping or saturation logic -- as part of
the classical control prior, not as a learned component.

\subsection{Reward Decomposition}

The episode reward sums seven per-step terms:
\begin{equation}
R = \sum_{t=1}^{T} \left( r^{\rm prog}_t + r^{\rm dist}_t + r^{\rm reach}_t + r^{\rm closest}_t + r^{\rm jerk}_t + r^{\rm align}_t + r^{\rm idle}_t \right).
\label{eq:reward}
\end{equation}

Of these, the jerk term plays a central role in the failure
analysis of Section~\ref{ssec:probes}:
\begin{equation}
r^{\rm jerk}_t = -\lambda_j \cdot \|\mathbf{a}_t - \mathbf{a}_{t-1}\| / dt,
\label{eq:jerk}
\end{equation}
with $\lambda_j\!=\!0.10$. Its global minimum on a fixed trajectory
is $\mathbf{v}\!\equiv\!\text{const}$, which interacts adversarially
with the prior bias of the language model toward the digit token
``0''.

\subsection{Evaluation Protocol}

Success criterion: final distance $<0.3$\,m and speed $<0.5$\,m/s
within the scenario time budget. Confidence intervals: Wilson 95\,\%
for SR, bootstrap 95\,\% (10\,000 resamples) for jerk. Pairwise
tests use Wilcoxon signed-rank with Holm--Bonferroni correction
across the 21 scenarios.

% =================================================================
\section{Experiments}
\subsection{Setup}

\textbf{Hardware.} A single consumer-grade NVIDIA RTX-class GPU
with 16\,GB VRAM, paired with a 16-thread x86 CPU and 64\,GB system
memory.

\textbf{Software.} PyTorch~2.4 with bfloat16, Hugging Face TRL~0.19
for GRPO, PEFT~0.11 for LoRA. The simulator is a custom NumPy/Python
implementation; no external robot simulator is required.

\textbf{Compute footprint.} The full experimental sweep totalled
approximately 5.4 GPU-hours, $\sim$1.6\,kWh of energy under typical
conditions. Reproducing the headline experiment (D-64 training
plus its 21-scenario evaluation) takes approximately 45 minutes on
the same hardware.

\textbf{Metrics.} Success rate (SR), mean trajectory jerk, and
training-time entropy are the three metrics that recur throughout
the results. SR is defined per-scenario and aggregated with Wilson
intervals; jerk is averaged over completed steps and aggregated
with bootstrap intervals.

% =================================================================
\section{Results}
\subsection{Quantitative Analysis: Vanilla GRPO Collapses}\label{ssec:collapse}

\begin{figure*}[!t]
\centering
\includegraphics[width=0.92\textwidth]{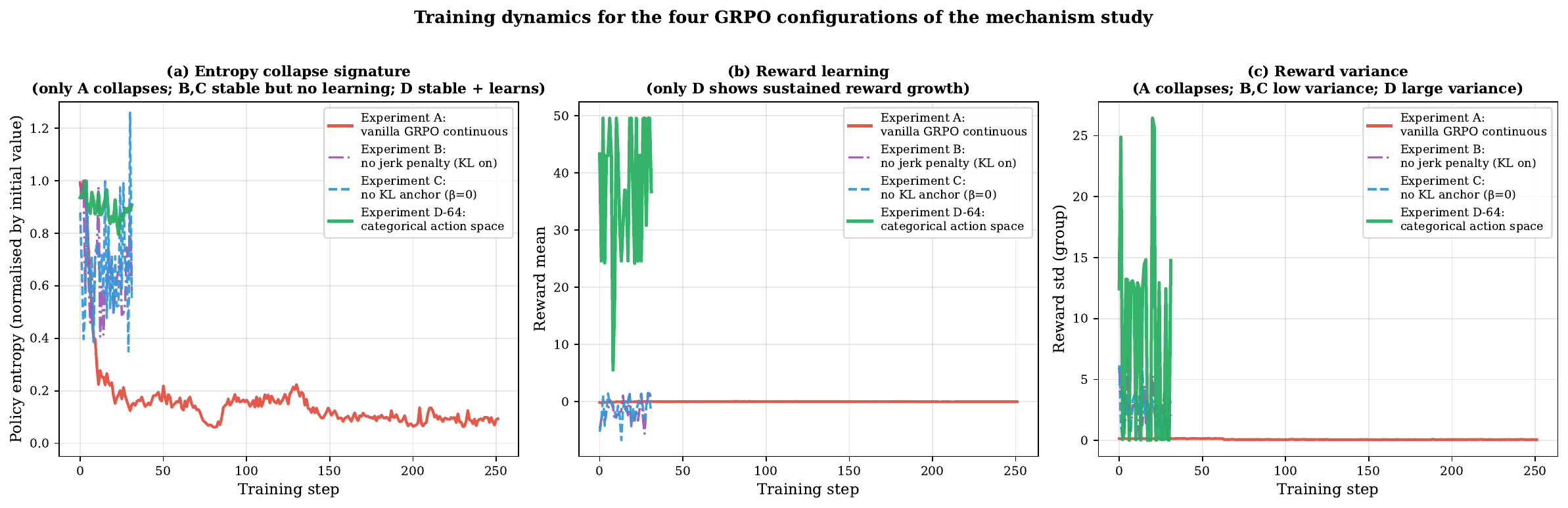}
\caption{Training dynamics for the four GRPO configurations of the
mechanism study. Entropy is normalised by the initial value to
make continuous-output runs (full LM vocabulary) and the discrete-
output run (5 labels) directly comparable. Only Experiment D
(categorical) both maintains entropy and grows reward. Experiments
B (no jerk penalty) and C (no KL anchor) prevent entropy collapse
but do not enable learning. Vanilla A collapses on both axes.}
\label{fig:training_dynamics}
\end{figure*}

Across 252 training steps the policy of Experiment A progressively
concentrates on the action $[0,0,0]$. Diagnostic signals
(Fig.~\ref{fig:training_dynamics}, panel a, red curve): policy
entropy $0.35\!\to\!0.03$ ($-91$\,\%); reward standard deviation
$0.15\!\to\!0.07$; mean completion length
$26.1\!\to\!21.4$ tokens. After step $\sim 60$, $\ge 99\,\%$ of
sampled completions begin with \texttt{"value":\,[0}. On
evaluation, the deployed policy emits byte-identical completions
across steps and scenarios; the drone neither moves under
no-disturbance scenarios nor resists wind under disturbance
scenarios. Aggregate: 0\,\% SR.

\subsection{Mechanism Probes}\label{ssec:probes}

\textit{1) Experiment B (no jerk penalty):} Final entropy stays
in the range 0.20--0.33 against 0.03 for vanilla. Reward bounces
between $-5$ and $+1.5$ across all 64 steps, with no upward trend.
Removing the jerk penalty avoids the entropy collapse but the
policy still does not find useful actions.

\textit{2) Experiment C ($\beta\!=\!0$):} The picture is almost
identical -- final entropy 0.20--0.30, reward in $[-1, +2]$.
Disabling the KL anchor produces the same outcome as B: collapse
avoided, learning not enabled.

The two probes therefore rule out the obvious one-factor
explanations. Neither the jerk-penalty trap nor the KL pull-back
to the prior is sufficient on its own to enable learning under a
continuous action space. We read this as evidence that the
remaining structural difference between Experiments B/C and
Experiment D -- the action-space type -- is doing the load-bearing
work; we do not claim this proves no other factor exists.

\subsection{Discrete Categorical Interface Restores Convergence}\label{ssec:tradeoff}

Experiment D (categorical, otherwise identical to A) restores
convergence within 64 GRPO steps
(Fig.~\ref{fig:training_dynamics}, green). The resulting
controller, which we call \textit{Strategy-v3}, exhibits a Pareto
trade-off between trajectory smoothness and full reliability along
training duration; we report both endpoints
(Table~\ref{tab:tradeoff}).

\begin{table}[!t]
\caption{Strategy-v3 trade-off endpoints.}
\label{tab:tradeoff}
\centering
\renewcommand{\arraystretch}{1.1}
\footnotesize
\begin{tabular}{lcc}
\toprule
& \textbf{D-64} & \textbf{D-256} \\
\midrule
GRPO training steps & 64 & 256 \\
Success rate (\%) & 98.6 & 100 \\
SR 95\,\% Wilson CI & [95.7,99.6] & [98.2,100] \\
SR-failure scenarios & 2 wind-extreme & none \\
Mean jerk (m/s$^3$) & \textbf{0.656} & 1.103 \\
Jerk 95\,\% bootstrap CI & [0.602,0.711] & - \\
Practitioner choice & smoothness & reliability \\
\bottomrule
\end{tabular}
\end{table}

The 1.4-pp reliability shortfall of D-64 is concentrated on
extreme-wind scenarios; D-256 closes that gap but at a measurable
jerk regression. Practitioners selecting along this trade-off are
choosing between ``smoother trajectories with rare extreme-wind
failures'' and ``always-feasible flight at higher jerk.''

\subsection{Classical Baseline and the Velocity-Cap Confound}\label{ssec:pid_baseline}

A single PID with $K_i\!=\!0.30$ and $v_{\max}\!=\!2.5$ matches the
100\,\% SR of D-256 with mean jerk 0.736\,m/s$^3$
(Table~\ref{tab:e7}). The categorical interface - not the
language model itself - is the architectural element responsible
for the reliability of Strategy-v3; the LM contributes (i) a path
to natural-language mission specification, (ii) smoothness benefits
at the under-trained operating point D-64, and (iii) compatibility
with hierarchical mission planners.

\begin{table}[!t]
\caption{PID $K_i$ and $v_{\max}$ sweep ($N\!=\!5$).}
\label{tab:e7}
\centering
\renewcommand{\arraystretch}{1.1}
\footnotesize
\begin{tabular}{lcc}
\toprule
PID variant & SR (\%) & Jerk (m/s$^3$) \\
\midrule
$K_i\!=\!0.05,v_{\max}\!=\!3.5$ (default) & 62.9 & 0.853 \\
$K_i\!=\!0.10,v_{\max}\!=\!3.5$ & 78.1 & 0.908 \\
$K_i\!=\!0.20,v_{\max}\!=\!3.5$ & 93.3 & 1.012 \\
$K_i\!=\!0.30,v_{\max}\!=\!3.5$ & 100.0 & 1.106 \\
$K_i\!=\!0.30,v_{\max}\!=\!2.5$ & 100.0 & \textbf{0.736} \\
\bottomrule
\end{tabular}
\end{table}

\textit{The $v_{\max}$ confound and its resolution.} Comparing the
bottom two rows of Table~\ref{tab:e7} reveals that approximately
one third of the smoothness benefit of the
$K_i\!=\!0.30,v_{\max}\!=\!2.5$ variant comes from velocity
saturation rather than integral action. We re-evaluated D-256 with
its output globally clipped to $v_{\max}\!=\!2.5$
(Table~\ref{tab:vmax_clipped}). Result: SR retained at 100\,\%,
mean jerk \textbf{0.796}\,m/s$^3$ (95\,\% bootstrap CI
[0.761,0.831]). The smoothness gap relative to PID-anti\_wind
closes from 50\,\% to 8\,\%.

\begin{table}[!t]
\caption{Apples-to-apples Strategy-v3-256 with output globally
clipped at $v_{\max}\!=\!2.5$.}
\label{tab:vmax_clipped}
\centering
\renewcommand{\arraystretch}{1.1}
\footnotesize
\begin{tabular}{lcc}
\toprule
Controller & SR (\%) & Jerk (m/s$^3$) \\
\midrule
Strategy-v3-256, $v_{\max}\!=\!3.5$ (orig.) & 100 & 1.103 \\
\textbf{Strategy-v3-256, $v_{\max}\!=\!2.5$} & \textbf{100} & \textbf{0.796} \\
PID-anti\_wind, $v_{\max}\!=\!2.5$ & 100 & 0.736 \\
\bottomrule
\end{tabular}
\end{table}

\subsection{Aggregate Eight-Controller Comparison}

\begin{figure*}[!t]
\centering
\includegraphics[width=0.92\textwidth]{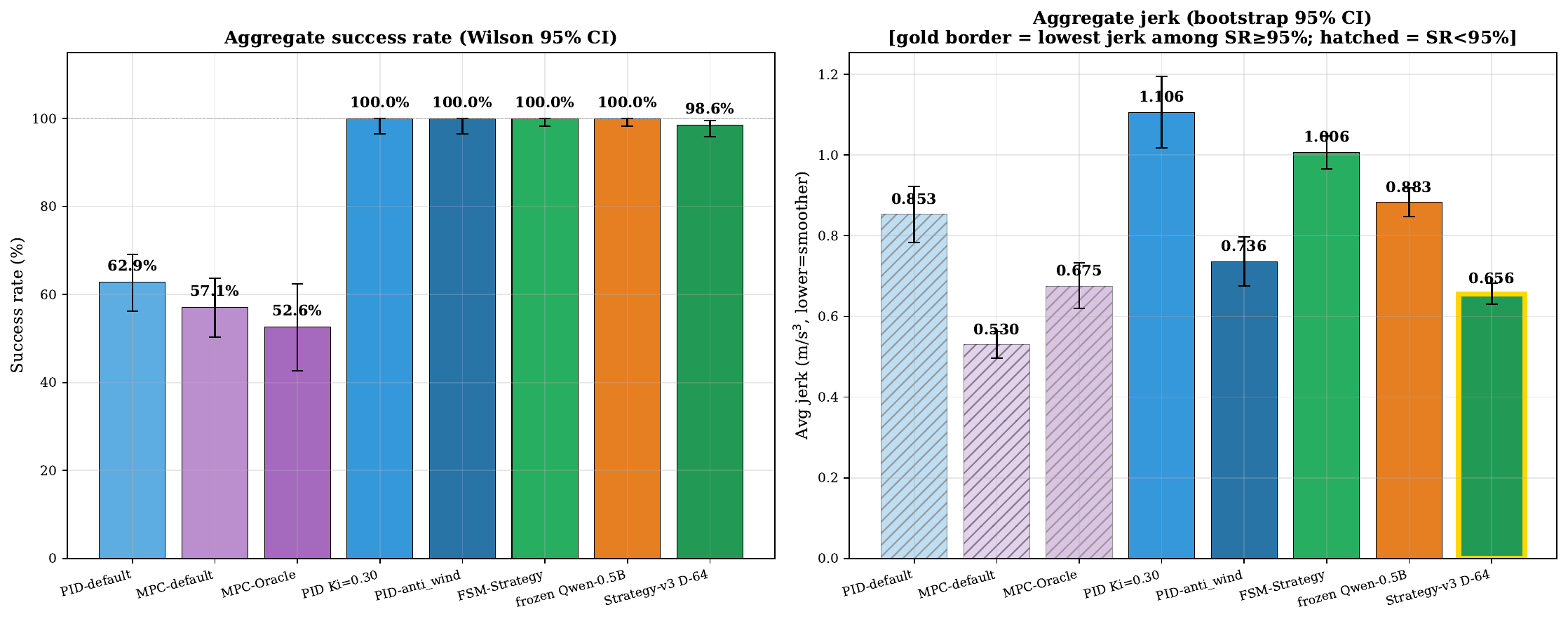}
\caption{Aggregate SR (Wilson 95\,\% CI) and jerk (bootstrap 95\,\%
CI) for eight controllers across 21 scenarios with $N\!=\!10$
seeds. The right-most green bar is the smoothness-priority
endpoint of the trained controller (D-64); D-256 is reported in
Tables~\ref{tab:tradeoff} and \ref{tab:vmax_clipped}.}
\label{fig:aggregate}
\end{figure*}

Fig.~\ref{fig:aggregate} aggregates SR and jerk across all eight
controllers we evaluated, with 95\,\% CIs. PID-default and the two
MPC variants fail (hatched bars, SR\,$<$\,95\,\%); the remaining
six controllers reach 100\,\% SR with separable jerk
distributions.

\subsection{Cross-Model Replication}\label{ssec:cross_model}

We swap the base model for SmolLM2-360M and SmolLM2-1.7B with
identical prompting and the same 5-way categorical interface. All
three reach 100\,\% SR on a 12-scenario subset with frozen weights
(Table~\ref{tab:cross_model}). The categorical-interface
contribution generalises across at least three LM families
spanning a 5$\times$ parameter range.

\begin{table}[!t]
\caption{Cross-model evaluation, frozen weights, $N\!=\!3$.}
\label{tab:cross_model}
\centering
\renewcommand{\arraystretch}{1.1}
\footnotesize
\begin{tabular}{lccc}
\toprule
Model & Params & SR (\%) & Jerk (m/s$^3$) \\
\midrule
Qwen2.5-0.5B & 0.49\,B & 100.0 & 0.873 \\
SmolLM2-360M & 0.36\,B & 100.0 & 1.096 \\
SmolLM2-1.7B & 1.71\,B & 100.0 & 1.148 \\
\bottomrule
\end{tabular}
\end{table}

\subsection{Bandwidth Sweep}\label{ssec:bandwidth}

\begin{figure}[!t]
\centering
\includegraphics[width=\columnwidth]{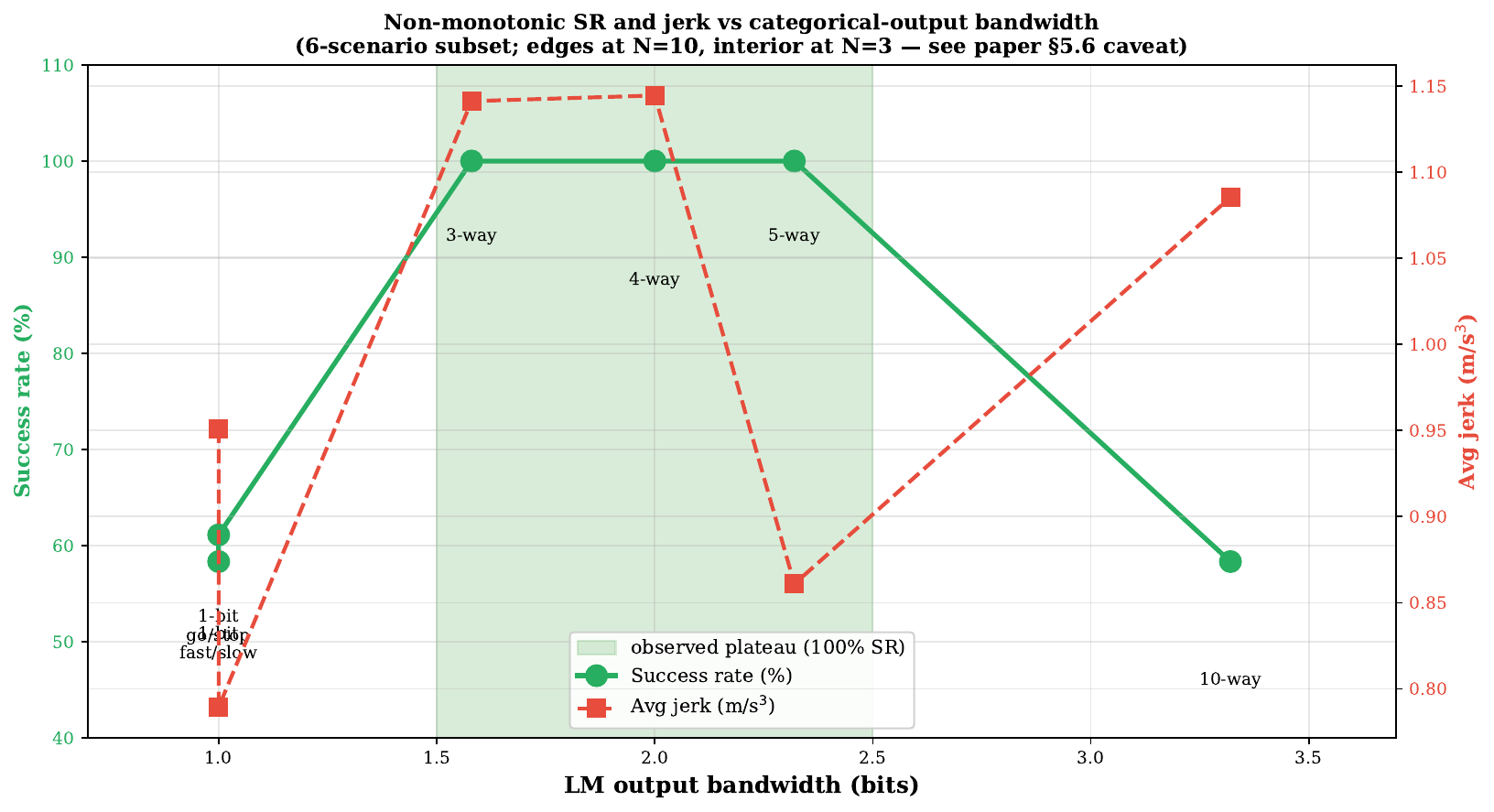}
\caption{Non-monotonic SR and jerk vs. categorical-output bandwidth
on a 6-scenario subset. The 1.58--2.32-bit range (green band)
shows an observed plateau at 100\,\% SR; both edges drop to
$\le 80\,\%$. Edge points at $N\!=\!10$; interior at $N\!=\!3$.}
\label{fig:bandwidth}
\end{figure}

Sweeping the LM output bandwidth from 1\,bit (\textsf{go/stop}) to
3.32\,bits (10-way categorical) on a 6-scenario subset, we observe
a non-monotonic relationship between bandwidth and reliability
(Fig.~\ref{fig:bandwidth}). 1.58--2.32\,bits achieves 100\,\% SR;
the edges drop to 58--80\,\%. Edge points evaluated at $N\!=\!10$;
interior points at $N\!=\!3$. Under a binomial model with
$N\!=\!3$ and observed $3/3 = 100\,\%$ per cell, the one-sided
Wilson 95\,\% lower bound on per-cell SR is $\sim 29\,\%$; that is,
the interior plateau is consistent with true per-cell SR
$\ge 90\,\%$ but does not rule out modest variation. We do not
claim 1.58--2.32\,bits as a fundamental constant; the range is
task-specific and we offer this as an empirical observation, not a
law.

% =================================================================
\section{Discussion}\label{sec:discussion}
\subsection{Why the Discrete Interface Works}

A categorical output over five labels assigns zero probability to
the literal action $[0,0,0]$, simply because no such token exists
in the alphabet. The prior bias on digit tokens, which dominates
the early gradients of vanilla GRPO, has no surface to act on.
Quantisation also collapses the parse-error problem: any sample
the LM produces is either a valid label or a parse failure that
falls back to a default; ``a wrong number'' becomes ``a still-valid
label''. We treat this as a necessary condition for learning in
our experiments; sufficient conditions in other settings are not
something Experiment D can establish alone.

The non-monotonic bandwidth curve of
Section~\ref{ssec:bandwidth} is consistent with two opposing
pressures on the softmax. A 1-bit \textsf{go/stop} alphabet is
too small to encode integral action against sustained disturbance
and forces the policy into one of two unsuitable modes. A 10-way
alphabet is too large: the softmax becomes noisier, regime
boundaries blur, and the rate of mis-classification rises. Whether
the empirical plateau between these two regimes generalises to
other tasks is open.

% =================================================================
\section{High-Fidelity Simulation Study}\label{sec:hwfid}

We do not deploy any controller to physical hardware in this
paper. The study in this section is purely a higher-fidelity
simulation -- not a hardware-in-the-loop nor a real-flight test
-- that augments our base simulator with the measurement and
actuation noise models of the Crazy{f}lie 2.1 platform:
Lighthouse-2 positioning noise (1.2\,cm RMS), IMU-integrated
velocity drift (5\,cm/s), and BLE-radio one-way delay (12\,ms).
Sensor and link parameters are taken from the Bitcraze
documentation~\cite{bitcraze} and the Greiff
thesis~\cite{greiff2017}. Five controllers were evaluated across
five Crazy{f}lie-feasible scenarios. PID-default reaches 100\,\%
success rate throughout; Strategy-v3-D-256 reaches 73\,\%, with
the hover scenario dropping to 0\,\% due to a training-distribution
mismatch (prompts contain no examples in which
$\|p\|\!\to\!0$ with persistent noise). The implied next experiment
is domain-randomisation training that includes hover regions with
realistic measurement noise; we do not attempt this in the present
paper.

% =================================================================
\section{Limitations}

\textbf{Single airframe class:} 1.5\,kg quadrotor only; behaviour
on heavier or smaller platforms is unknown.

\textbf{Single base-model size at training:} Strategy-v3 was
trained only on Qwen-0.5B. The cross-model frozen replication
(Section~\ref{ssec:cross_model}) shows the recipe transfers to
other backbones with frozen weights; the trained counterpart on
those backbones is open work.

\textbf{Single benchmark:} 21 scenarios in our simulator. Transfer
to other simulators and especially to real flight hardware is open.

\textbf{Reward hand-tuning:} The 7-term reward was designed
manually. Automated reward search via LM-assisted reward
auditing~\cite{eureka} might reveal additional collapse-avoidance
variants and is open work.

\textbf{Bandwidth-curve interior at $N\!=\!3$:} Disclosed binomial
CIs and discussed explicitly; full $N\!=\!10$ across all six
bandwidth points was not budget-feasible.

\textbf{$v_{\max}$ confound resolved post-hoc:} The cleanest
comparison point would be a single Strategy-v3 training with
$v_{\max}\!=\!2.5$ baked into all presets; we did not budget
compute for this.

% =================================================================
\section{Conclusion}

The headline finding of this paper is structural: GRPO converges
on small language models in continuous-control settings only when
the action space is discrete. Of the four ablations we ran, three
either collapsed the policy or stalled in noise; the
discrete-categorical configuration was the only one that
produced a learning signal. This places the failure of vanilla
GRPO outside the usual suspects of reward shaping or KL
regularisation, and inside the geometry of the output space.

Around this result sit four supporting findings. The trained
controller's smoothness--reliability behaviour is a Pareto curve,
not a single operating point, so we report both ends of it. The
categorical interface -- not the language model itself --
carries the reliability; a re-tuned PID matches the trained
controller's success rate without the LM. The recipe transfers
across three pretrained LM families, confirming it is not
Qwen-specific. The high-fidelity simulation surfaces a hover-region
training-distribution gap that points squarely at the next
experimental step.

Real-hardware deployment remains future work, as does training
under deliberate domain randomisation in the hover regime
identified above.

% =================================================================
\bibliographystyle{IEEEtran}

\end{document}